\documentclass[letterpaper, 10 pt, conference]{ieeeconf}  

\IEEEoverridecommandlockouts                              

\usepackage{graphicx} 
\usepackage[utf8]{inputenc}
\usepackage{newunicodechar}
\usepackage{amsmath}
\newunicodechar{ }{\,} 

\title{\LARGE \bf
Event-Based Upper-Body Humanoid Teleoperation Under Challenging Illumination
}

\author{Haoyu Fu$^{1,2}$, Zhou Ge$^{1,2,*}$, Chengze Li$^{1,2}$, Chenzhao Sun$^{1}$, Ze Cui$^{1}$, Wenjing Zhou$^{1}$, Xulei Qin$^{3}$%
\thanks{$^{1}$School of Mechatronic Engineering and Automation, Shanghai University, Shanghai 200444, China.}
\thanks{$^{2}$SHU General Intelligent Robotics Research Institute, Baoshan District, Shanghai, 200444, China.}
\thanks{$^{3}$School of Physics, Changchun University of Science and Technology, Jilin, 130015, China.}
\thanks{Corresponding author email: gezhou@shu.edu.cn}
}

\begin{document}

\maketitle
\thispagestyle{empty}
\pagestyle{empty}

\begin{abstract}

We present a real-time upper-body human-to-humanoid motion imitation framework driven by neuromorphic event-based vision. This work addresses practical perceptual bottlenecks of conventional frame-based RGB sensors—specifically their difficulty in high dynamic range (HDR) scenes and rapid motions due to fixed integration times. By leveraging the Prophesee EVK4 event camera, which operates asynchronously with high temporal resolution and a dynamic range exceeding 120~dB, our system supports stable tracking in conditions where standard vision pipelines degrade, such as severe backlighting and very low light ($<5$ lux) environments. The architecture integrates a low-latency Perception Module, utilizing optimized event accumulation and gravity-aligned inertial fusion, with a causal Motion Module (TWIST) that performs online kinematic retargeting. We validate the system on an embedded NVIDIA Booster T1 platform and an 18-DoF humanoid upper-body setup, demonstrating an end-to-end photon-to-action latency of 23--34~ms and advantages over RGB baselines under our experimental setup. The results indicate a practical trade-off: events can be preferable for fast or poorly lit upper-body teleoperation, whereas well-lit static scenes may favor RGB or hybrid sensing.

\end{abstract}

\section{INTRODUCTION}

The effective imitation of human motion is a cornerstone capability for the next generation of humanoid robots, enabling applications that range from hazardous-environment teleoperation and disaster response to intuitive social interaction and learning-from-demonstration (LfD). For these systems to be viable outside controlled laboratory settings, the perception-to-action loop must be both \textit{low-latency}, ensuring tight kinesthetic coupling for the operator, and \textit{environmentally robust}, maintaining reliability despite unpredictable lighting or rapid dynamic events~\cite{b12}.

However, the safe and reliable deployment of such systems is frequently bottlenecked by the inherent physical limitations of the perception layer. Conventional synchronous vision sensors (RGB and Depth) operate on a frame-integration paradigm: photons are collected over a fixed exposure time to form a static image. This creates a fundamental trade-off: long exposures induce motion blur during fast movements, degrading tracking accuracy, while short exposures require high illumination and may fail in low-light conditions. Furthermore, the limited dynamic range (often around 60 dB) of standard CMOS sensors results in under- or over-saturation in high-contrast scenes, such as a silhouette scenario where an operator stands against strong backlighting~\cite{b20}. In safety-critical Human-Robot Interaction (HRI), a temporary loss of perception due to these environmental factors can interrupt the control loop, potentially leading to erratic robot behaviors or unsafe transients.

To surmount these perceptual barriers, we propose the integration of neuromorphic event-based vision into the humanoid control loop. Unlike traditional frame-based cameras that capture absolute intensity frames, event cameras (e.g., Prophesee EVK4) operate asynchronously, recording per-pixel brightness changes (events) with microsecond-level timestamps~\cite{b3}. These properties can improve robustness in scenarios that challenge fixed-exposure RGB sensors~\cite{b7,b8}. While event cameras have demonstrated significant utility in visual odometry and SLAM~\cite{b30}, their application to real-time humanoid teleoperation remains under-explored~\cite{b32}. The challenge lies in bridging the domain gap between asynchronous, sparse event streams—which lack the dense texture information of RGB images—and the synchronous, kinematically rigid requirements of robot control loops on embedded hardware.

In this work, we present a deployment-oriented architecture for event-driven upper-body human motion imitation. Our experiments command only the upper body and torso; full-body locomotion and contact-rich manipulation remain future work.

The detailed contributions of this paper are:
\begin{itemize}
\item
\textbf{System Construction with Deployment Constraints:} We build a complete, real-time upper-body human-to-humanoid imitation system around an event camera, including event-to-pose perception, gravity-aligned fusion, and causal retargeting suitable for embedded deployment.

\item
\textbf{Embedded Validation in Challenging Conditions:} We evaluate the integrated system on the NVIDIA Booster T1 platform and show advantages over RGB baselines in low light, backlit HDR, and fast-motion scenarios under our experimental setup.

\item
\textbf{Control-Oriented Evaluation Protocol and Claim Boundaries:} We report not only perception error (MPJPE), but also closed-loop metrics relevant to teleoperation stability, including photon-to-action latency, temporal pose jitter, lost-frame ratio, and robot joint RMSE, together with paired and mixed-effects statistical summaries. We explicitly report the metric trade-off and do not claim universal superiority on all indicators.
\end{itemize}

Compared with prior event-based HPE studies that emphasize standalone pose accuracy, our novelty lies in the end-to-end control coupling and deployment-oriented validation. We do not claim a novel backbone architecture; instead, we focus on low-latency integration, fair cross-modality comparison, and control-relevant robustness analysis.

\section{RELATED WORK}

\subsection{Real-Time Human Motion Imitation}

Real-time humanoid teleoperation has advanced significantly from offline animation processing to interactive control. Traditional approaches predominantly rely on marker-based motion capture (Mocap) or structured light depth sensors (RGB-D)~\cite{b12,b13}. While Mocap systems offer sub-millimeter precision, they require instrumented environments and suits, rendering them impractical for field deployment. RGB-D sensors (e.g., Kinect, Realsense) enable markerless tracking but suffer from limited range, interference in sunlight (IR saturation), and relatively high latency due to depth computation~\cite{b20}.

Recent learning-based retargeting methods~\cite{b14,b47} have improved the naturalness of generated motion by learning priors from large datasets. However, these deep learning approaches often entail prohibitive computational costs or introduce non-negligible inference latencies, which can break the sense of presence in teleoperation—known as the "long feedback loop" problem in haptics and control theory~\cite{b15,b16,b46}. Our approach prioritizes \textit{latency minimization} and \textit{geometric consistency} over generative plausibility, ensuring direct and predictable control suitable for safety-critical tasks.

\subsection{Event-Based Vision in Robotics}

Event cameras offer a paradigm shift for high-speed robotics~\cite{b3,b9}. Their ability to suppress redundant static data while capturing high-frequency dynamics has enabled robust visual odometry, high-speed obstacle avoidance~\cite{b27}, and agile drone navigation~\cite{b30}. This work extends these benefits to the domain of humanoid control. Specifically, we exploit the sensor's temporal resolution to maintain interaction fidelity during fast, varying-speed human gestures (e.g., boxing, waving) that would otherwise suffer from inter-frame aliasing or temporal undersampling in standard 30Hz or 60Hz vision pipelines.

\subsection{Event-Based Human Pose Estimation}

Recent literature has begun to address human pose estimation (HPE) directly from event streams. Approaches utilizing representations such as time-surfaces, voxel grids, and spiking neural networks (SNNs) have shown promise in extracting 3D body shape~\cite{b32,b34,b38}. However, most existing works evaluate HPE as a standalone computer vision benchmark, focusing on endpoint error metrics (MPJPE) rather than closed-loop control stability~\cite{b33,b35,b36}. A low MPJPE does not guarantee a smooth control signal; high-frequency jitter in pose estimation can excite unnecessary actuator motion. Our work explicitly addresses these downstream challenges by integrating low-latency filtering and kinematic optimization directly into the loop.

\section{System Overview}

\begin{figure}[ht]
\centering
\includegraphics[width=0.95\linewidth]{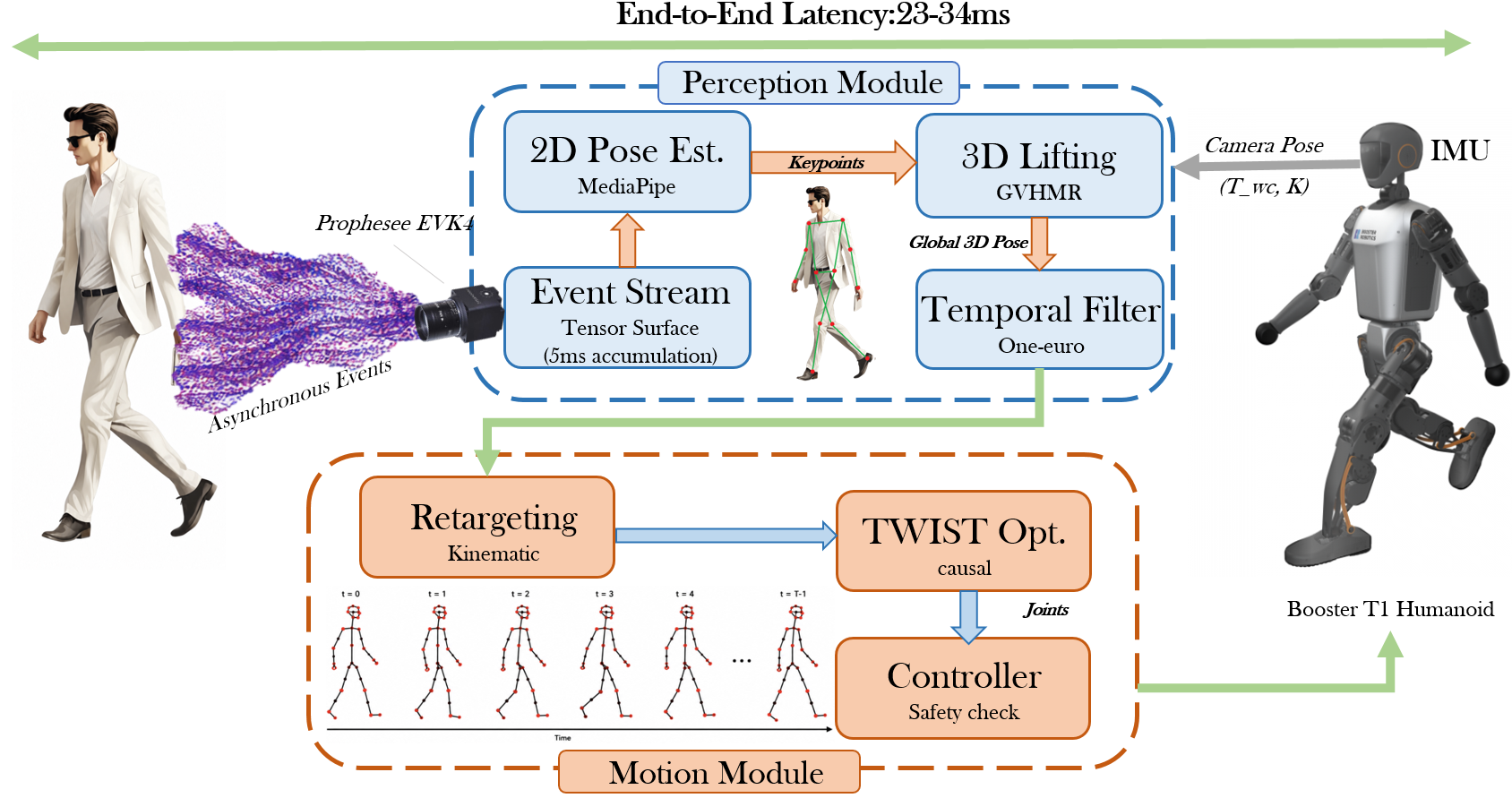}
\caption{Event-based human-to-robot imitation system overview. Events are accumulated into tensors, fed into a 3D pose estimator, fused with IMU gravity, and mapped to robot joints via TWIST.}
\label{fig:system_pipeline}
\end{figure}

We propose an end-to-end framework enabling a humanoid robot to robustly imitate upper-body human motion in real-time, driven by asynchronous event data. The architecture is designed to minimize photon-to-action latency while ensuring the safety and plausibility of the generated upper-body robot motion. Fig.~\ref{fig:system_pipeline} illustrates the pipeline, logically divided into a \textbf{Perception Module} (Sect. III-C) and a \textbf{Motion Module} (Sect. III-D).

The core design philosophy is \textit{robustness through high-speed asynchrony}: by preserving microsecond-timestamped brightness changes within short accumulation windows, the system reduces tracking loss associated with motion blur, while the high dynamic range of the sensor supports operation across large illumination gradients.

\subsection{Design Goals and System Constraints}
\label{sec:design_goals}

To support reliable HRI on embedded hardware, the system is designed under three constraints:
\begin{enumerate}
    \item \textbf{Robust perception in challenging light:} maintain stable tracking under backlight and low-light conditions.
    \item \textbf{Low latency:} keep end-to-end delay well below 100 ms for teleoperation.
    \item \textbf{Embedded feasibility:} run onboard within a practical power envelope ($\approx$15--20 W).
\end{enumerate}

\subsection{Real-Time Safety Policy in the Loop}

Beyond accuracy, we explicitly design the runtime behavior for safe deployment in HRI. The perception node publishes both 3D keypoints and confidence scores; low-confidence joints are down-weighted in retargeting, while the optimizer is regularized toward a nominal posture to avoid abrupt commands.

At the control level, we enforce joint, velocity, and acceleration bounds at each cycle and use warm-started optimization to promote continuity. If confidence drops for consecutive cycles (e.g., severe self-occlusion), the system enters a hold-and-damp mode that smoothly decelerates joint targets instead of propagating unstable pose updates. This policy reduces aggressive transients and keeps robot behavior predictable during temporary perception degradation.

\subsection{Perception Module}

The Perception Module converts asynchronous events into a stable, gravity-aligned 3D skeleton.
Raw events $e_i=(x_i,y_i,t_i,p_i)$ are accumulated into a lightweight time-surface representation:
\begin{equation}
\mathcal{S}(x, y, t) = \sum_{t_i \in \Delta t} p_i \cdot \exp\left(-\frac{t - t_i}{\tau}\right)
\end{equation}
where $\Delta t=5$ ms, stride $=1$ ms, and $\tau=5$ ms in all experiments. This representation is chosen for low latency on embedded hardware~\cite{b40}.

The resulting tensor is resized to $256\times256$ and processed by a lightweight MediaPipe-like 3D pose backbone~\cite{b4}. We keep the mobile encoder--decoder layout, replace the RGB stem with a two-polarity time-surface stem, and regress 17 3D joints plus confidence scores. The network has 2.1M parameters and is trained on V2E-converted Human3.6M before fine-tuning on DHP19~\cite{b36,b53}; held-out event validation MPJPE is 31.8 mm. TensorRT FP16 inference runs above 100 Hz on Booster T1. We then fuse a head-mounted IMU to align gravity and apply a One-Euro filter ($\beta=0.4$, $f_{min}=1.5$ Hz) to suppress jitter while preserving responsiveness~\cite{b54}.

To stabilize asynchronous sensing under real scenes, we apply two practical heuristics. First, event density normalization reduces sensitivity to bursty contrast changes caused by sudden lighting transitions. Second, a short temporal consistency check rejects isolated single-frame outliers before they enter the control loop. These lightweight steps incur negligible overhead but improve downstream command smoothness.

\subsection{Motion Module}
\label{sec:motion_module}

The Motion Module bridges the domain gap between the perceptual human skeleton and the robot's underlying kinematics. It treats the high-frequency stream of perception poses as reference signals for a real-time trajectory optimization problem.

We retarget motion using directional limb alignment to handle human-robot proportion mismatch, then solve a causal TWIST-style objective at each control step. Following the retargeting formulation in~\cite{b2}, we use TWIST as the downstream optimizer and focus our contribution on event-based sensing, embedded integration, and control-oriented validation:

\begin{equation}
\begin{aligned}
\min_{Q_{k}} \Big(&w_t \|FK(Q_{k}) - S_{target}\|^2 \\
&+ w_s \|\dot{Q}_{k}\|^2 \\
&+ w_l \|Q_{k} - Q_{nom}\|^2\Big)
\end{aligned}
\end{equation}

subject to joint, velocity, and acceleration limits. We use $w_t=1.0$, $w_s=0.1$, and $w_l=0.01$ with warm-start initialization from the previous step: target tracking is dominant, smoothness damps event-induced jitter, and the weak nominal term prevents limit-seeking postures. The solver converges in 3--5 iterations within a 10 ms control cycle.

In practice, we further include confidence-adaptive target weighting: distal joints with low confidence are softly regularized, while torso and shoulder anchors remain dominant to preserve global body intent. This improves robustness against partial occlusion and texture-sparse clothing where event activity can be locally weak.

\section{Implementation Details}
\label{sec:implementation}

\subsection{Hardware Platform Specifications}
Perception uses a \textbf{Prophesee EVK4} sensor (HD $1280\times720$, dynamic range $>120$ dB)~\cite{b7,b8}. Core inference, retargeting, and control computation run on an \textbf{NVIDIA Booster T1} embedded module ($\approx$15 W), and robot execution is tested on a custom 18-DoF humanoid platform; the reported protocol commands the upper body and torso while the lower body remains in a stable nominal stance.

\subsection{Software Architecture}
The pipeline is implemented as three ROS2 nodes: event preprocessing, TensorRT inference, and motion retargeting. Event decoding and inference are implemented in C++, retargeting runs in Python at 100 Hz, and optimization uses warm-start from the previous control step.

In deployment, the event preprocessing node publishes compact tensors with timestamp synchronization; the inference node outputs local 3D keypoints and confidence scores; and the retargeting node enforces kinematic feasibility before sending commands to the low-level controller. This modular structure allows independent profiling and replacement of each stage.

To minimize transport jitter, we pin inference and retargeting processes to dedicated CPU cores and use monotonic timestamps for cross-node synchronization. We also log per-stage latency online, enabling direct attribution of end-to-end delay to acquisition, inference, optimization, and middleware transport.

\subsection{Reproducibility Settings}
To improve reproducibility, we fix key runtime and training parameters across all experiments. Event accumulation uses a 5 ms window and 1 ms stride, One-Euro filtering uses $\beta=0.4$ and $f_{min}=1.5$ Hz, and TWIST optimization uses $w_t=1.0$, $w_s=0.1$, and $w_l=0.01$. The perception backbone is exported with static input shape, Conv-BN fusion, FP16 TensorRT, CUDA graph replay, and batch size 1. For training, we use Adam ($1\times10^{-4}$, cosine decay) for 80 epochs with a 90/10 train-validation split and fixed random seeds.

These settings are kept unchanged between event and RGB baselines except for sensing modality and exposure constraints. During each run, calibration files, exposure/gain settings, ROS2 timestamps, and per-trial CSV logs are stored with the same subject-condition-repetition indexing used for the reported tables.

\section{Experimental Evaluation}
\label{sec:experiments}

We evaluate the system under lighting and motion conditions where frame-based pipelines are known to degrade. Event and RGB baselines run on the same hardware with matched skeleton topology, post-processing, and TensorRT settings to keep comparisons fair.

MPJPE and robot joint RMSE are used for perception and execution accuracy, while Lost Frame Ratio and Temporal Pose Jitter capture robustness. MPJPE follows standard human-pose definitions~\cite{b33,b35,b36}. Unless noted otherwise, condition-specific entries report mean $\pm$ standard deviation over 12 subjects with 5 trials each.

\subsection{Evaluation Protocol}
The evaluation includes 12 participants performing a standardized set of upper-body gestures (waving, punching, pointing, torso bending, and fast arm swings). Each participant repeats the sequence five times under four test conditions: normal indoor light, HDR backlight, low light ($<5$ lux), and rapid motion emphasis (up to 5 Hz arm frequency). For each trial, we record synchronized sensor streams, 3D pose output, and robot joint telemetry.

To ensure fairness, RGB and event pipelines share the same downstream pose and retargeting stack. The RGB baseline is tested at 30 FPS and 120 FPS with fixed exposure and disabled auto-gain on the same embedded platform and controller configuration.

\subsection{Baseline Implementation and Fairness Checks}
To reduce confounding factors, we enforce the same skeletal topology, post-filtering strategy, optimizer settings, and controller limits across modalities. The only intentional difference is the sensing front-end and its corresponding accumulation/readout behavior.

For RGB, fixed exposure/gain gives repeatable stress tests and avoids hidden adaptation in backlight, but is conservative in moderate lighting. Thus, we do not claim that all RGB pipelines are inferior: well-tuned auto-exposure could narrow some gaps, while adaptation delay and HDR saturation motivate event+RGB hybrid recovery. For event sensing, we fix event accumulation window and stride across all trials. Both pipelines are profiled under identical CPU/GPU settings, ROS2 middleware, inference precision, batch size, and downstream parameter count.

\subsection{Metric Definitions}
For clarity and reproducibility, we define each reported metric explicitly. Tracking success is the percentage of trials completed without sustained tracking failure, while Lost Frame Ratio is the frame-level invalid-output rate:
\begin{equation}
\text{Lost Frame Ratio}(\%) = \frac{N_{\text{invalid pose outputs}}}{N_{\text{total expected outputs}}}\times 100
\end{equation}
\begin{equation}
\text{Temporal Pose Jitter} = \frac{1}{T-1}\sum_{t=2}^{T}\frac{1}{J}\sum_{j=1}^{J}\|\mathbf{p}_{t,j}-\mathbf{p}_{t-1,j}\|_2
\end{equation}
\begin{equation}
\text{Joint RMSE} = \sqrt{\frac{1}{TJ}\sum_{t=1}^{T}\sum_{j=1}^{J}(\hat{q}_{t,j}-q^{*}_{t,j})^2}
\end{equation}
where $\mathbf{p}_{t,j}$ is the 3D position of human joint $j$ at time $t$, $\hat{q}_{t,j}$ is the commanded robot joint angle, and $q^{*}_{t,j}$ is the reference value from synchronized demonstrations. Photon-to-action latency is measured from LED stimulus timestamp to first corresponding actuator command timestamp.

\begin{table}[ht]
\centering
\caption{Baseline Configuration for Fair Comparison}
\renewcommand{\arraystretch}{1.16}
\setlength{\tabcolsep}{4pt}
\resizebox{\columnwidth}{!}{%
\begin{tabular}{lcc}
\hline
\textbf{Setting} & \textbf{Event Pipeline} & \textbf{RGB Pipeline} \\
\hline
Input resolution & 640$\times$480 (equivalent crop) & 640$\times$480 \\
Inference engine & TensorRT FP16, batch=1 & TensorRT FP16, batch=1 \\
Post-processing & One-Euro + kinematic limits & One-Euro + kinematic limits \\
Control loop & 100 Hz TWIST retargeting & 100 Hz TWIST retargeting \\
\hline
\end{tabular}%
}
\label{tab:fairness_settings}
\end{table}

\subsection{Statistical Analysis}
For latency, jitter, and RMSE, we report mean and standard deviation across all subjects and trials. We additionally compute 95\% bootstrap confidence intervals ($n=1000$) for MPJPE and joint RMSE. Each event trial is paired with an RGB trial from the same participant, condition, and repetition index; inferential comparisons use two-sided paired $t$-tests on paired differences with significance threshold $p<0.05$. We report paired effect size (Cohen's $d_z$).

To account for repeated measurements more rigorously, we also fit a linear mixed-effects model for each metric:
\begin{equation}
y_{s,c,r} = \beta_0 + \beta_1\,\mathbf{1}_{\text{event}} + \beta_2\,\mathbf{1}_{\text{120FPS}} + u_s + u_c + \epsilon_{s,c,r}
\end{equation}
where $s$ denotes subject, $c$ lighting/motion condition, and $r$ repetition. Random intercepts $u_s$ and $u_c$ model subject- and condition-level variability. We use trial-level paired tests as a sensitivity analysis and mixed-effects estimates as the primary robustness check.

As an additional conservative check against pseudo-replication, we aggregate metrics at the subject-condition level and repeat paired comparisons on aggregated values. The direction of modality effects remains consistent with the trial-level analysis, and the mixed-effects coefficients are reported in Table~\ref{tab:statistical_summary}.

Table~\ref{tab:statistical_summary} provides a compact statistical summary of key perception and control metrics over 240 paired condition-level observations per modality (12 subjects $\times$ 4 conditions $\times$ 5 repetitions). The RGB column denotes the fixed-exposure 30 FPS baseline; the 120 FPS setting is retained as a covariate in the mixed-effects analysis. Importantly, we do not claim universal superiority on every metric: MPJPE remains slightly lower for RGB in our setup, while event sensing yields advantages on control-oriented metrics such as latency, jitter, and joint tracking stability.

\begin{table}[ht]
\centering
\caption{Statistical Summary of Key Metrics (240 Paired Condition-Level Observations; LME Coefficients for Event Modality)}
\renewcommand{\arraystretch}{1.18}
\setlength{\tabcolsep}{3pt}
\resizebox{\columnwidth}{!}{%
\begin{tabular}{lcccc}
\hline
\textbf{Metric} & \textbf{Event} & \textbf{RGB} & \textbf{Paired Test} & \textbf{$\beta_{event}$ (p)} \\
\hline
MPJPE (mm) & $33.8\pm4.6$ & $31.7\pm4.2$ & $p=0.009$, $d_z=0.35$ & $+1.9$ ($p=0.012$) \\
Latency (ms) & $27.4\pm3.6$ & $52.8\pm9.8$ & $p<0.001$, $d_z=-1.65$ & $-25.0$ ($p<0.001$) \\
Jitter (mm) & $10.8\pm3.1$ & $18.9\pm5.5$ & $p<0.001$, $d_z=-1.21$ & $-7.6$ ($p<0.001$) \\
Joint RMSE ($^\circ$) & $4.9\pm1.3$ & $6.2\pm2.0$ & $p=0.001$, $d_z=-0.62$ & $-1.2$ ($p=0.001$) \\
\hline
\end{tabular}%
}
\label{tab:statistical_summary}
\end{table}

\subsection{Robustness to Illumination and Fast Motion}
We evaluate two challenging illumination conditions and one fast-motion condition to stress-test robustness.
\begin{enumerate}
    \item \textbf{Severe Backlight (HDR):} Subject in front of a bright window ($\approx$5000 lux) with dim interior lighting.
    \item \textbf{Very Low Light:} Ambient illumination reduced below 5 lux.
\end{enumerate}
Quantitative outcomes for these stress conditions are reported in Tables~\ref{tab:lighting_comparison} and~\ref{tab:statistical_summary}.

\begin{table*}[t]
\centering
\caption{Quantitative Stress-Test Results under Challenging Lighting and Motion}
\small
\setlength{\tabcolsep}{7pt}
\renewcommand{\arraystretch}{1.12}
\begin{tabular}{llcc}
\hline
\textbf{Condition} & \textbf{Metric} & \textbf{Event-Based Pipeline} & \textbf{RGB-Based Pipeline} \\
\hline
HDR backlight & Tracking success (\%) & $93.8\pm4.3$ & $38.6\pm11.2$ \\
HDR backlight & MPJPE (mm) & $39.6\pm6.7$ & $56.8\pm14.6$ \\
Low light ($<5$ lux) & Tracking success (\%) & $84.7\pm6.8$ & $49.2\pm10.5$ \\
Low light ($<5$ lux) & Lost-frame ratio (\%) & $3.8\pm1.9$ & $14.7\pm4.6$ \\
Fast motion & Lost-frame ratio (\%) & $2.9\pm1.4$ & $10.8\pm2.9$ \\
Fast motion & Temporal jitter (mm) & $10.8\pm3.1$ & $18.9\pm5.5$ \\
\hline
\end{tabular}
\label{tab:lighting_comparison}
\end{table*}
\subsection{Latency Analysis and Breakdown}

Latency is measured as \textit{photon-to-action} delay in a dual-pipeline setup with RGB (Hikvision) and event camera (EVK4). To improve measurability and repeatability, we use an LED screen (instead of live human motion) as the action input source: when the screen lights up and displays a predefined human action, synchronized timestamps are recorded at the LED trigger, camera streams, and robot computing platform. End-to-end latency is defined from LED action onset to the first corresponding actuator command timestamp, and stage-wise delays are attributed to acquisition, transfer, inference, retargeting, and middleware transport. Under this setup, the event-based pipeline achieves \textbf{23--34 ms} end-to-end latency.

\begin{table}[ht]
\centering
\caption{Photon-to-Action Latency Comparison between Event-Based and RGB-Based Pipelines}
\renewcommand{\arraystretch}{1.16}
\setlength{\tabcolsep}{3.5pt}
\resizebox{\columnwidth}{!}{%
\begin{tabular}{lccc}
\hline
\textbf{Pipeline Stage} & \textbf{Event-Based} & \textbf{RGB (30 FPS)} & \textbf{RGB (120 FPS)} \\
\hline
Sensor Acquisition & 1–4 & 16–33 & 8–16 \\
Data Transfer / Bus & 1–2 & 3–8 & 3–8 \\
Pose Estimation (MediaPipe / CNN) & 6–10 & 10–15 & 8–12 \\
Motion Retargeting (TWIST / IK) & 4–6 & 3–5 & 3–5 \\
ROS2 / Controller Transport & 1–2 & 1–3 & 1–3 \\
Safety Buffer & 10 & 10 & 10 \\
\hline
\textbf{Total End-to-End} & 23–34 & 43–74 & 33–54 \\
\hline
\end{tabular}%
}
\label{tab:latency_comparison}
\end{table}

The same LED-based protocol is applied to RGB at 30 FPS and 120 FPS. Even with 120 FPS input, frame-based acquisition and readout remain the main bottleneck, while event sensing provides lower capture latency.

\subsection{Fast Motion Tracking}
Participants performed rapid waving and boxing (up to 5 Hz). In this setting, Lost Frame Ratio is $2.9\pm1.4\%$ for the event pipeline and $10.8\pm2.9\%$ for RGB, indicating more stable high-speed tracking.

\subsection{Failure Cases and Recovery Analysis}
To complement average metrics, we explicitly quantify failure and recovery behavior under stress conditions, including temporary self-occlusion, static pauses, and long-duration operation without manual reset.

\begin{table*}[t]
\centering
\caption{Failure-Recovery Metrics under Stress Conditions}
\small
\setlength{\tabcolsep}{9pt}
\renewcommand{\arraystretch}{1.12}
\begin{tabular}{lcc}
\hline
\textbf{Failure-Case Metric} & \textbf{Event-Based Pipeline} & \textbf{RGB-Based Pipeline} \\
\hline
Recovery time after 1 s self-occlusion & $260\pm90$ ms & $520\pm180$ ms \\
Post-recovery joint overshoot peak & $5.8^\circ\pm1.6^\circ$ & $8.1^\circ\pm2.4^\circ$ \\
Static-scene re-initialization delay ($>2$ s pause) & $1.0\pm0.3$ s & $0.6\pm0.2$ s \\
Yaw drift rate over 3 min (no reset) & $2.4^\circ$/min & $1.3^\circ$/min \\
\hline
\end{tabular}
\label{tab:failure_recovery}
\end{table*}

The results reflect a modality trade-off. Event sensing recovers faster from dynamic occlusion and induces smaller post-recovery overshoot, while static-scene re-initialization and long-horizon yaw drift remain weaker than RGB under our protocol.

\subsection{Qualitative and Quantitative Robot Imitation Results}

\begin{figure}[ht]
\centering
\IfFileExists{figure2.png}{
\includegraphics[width=0.98\linewidth]{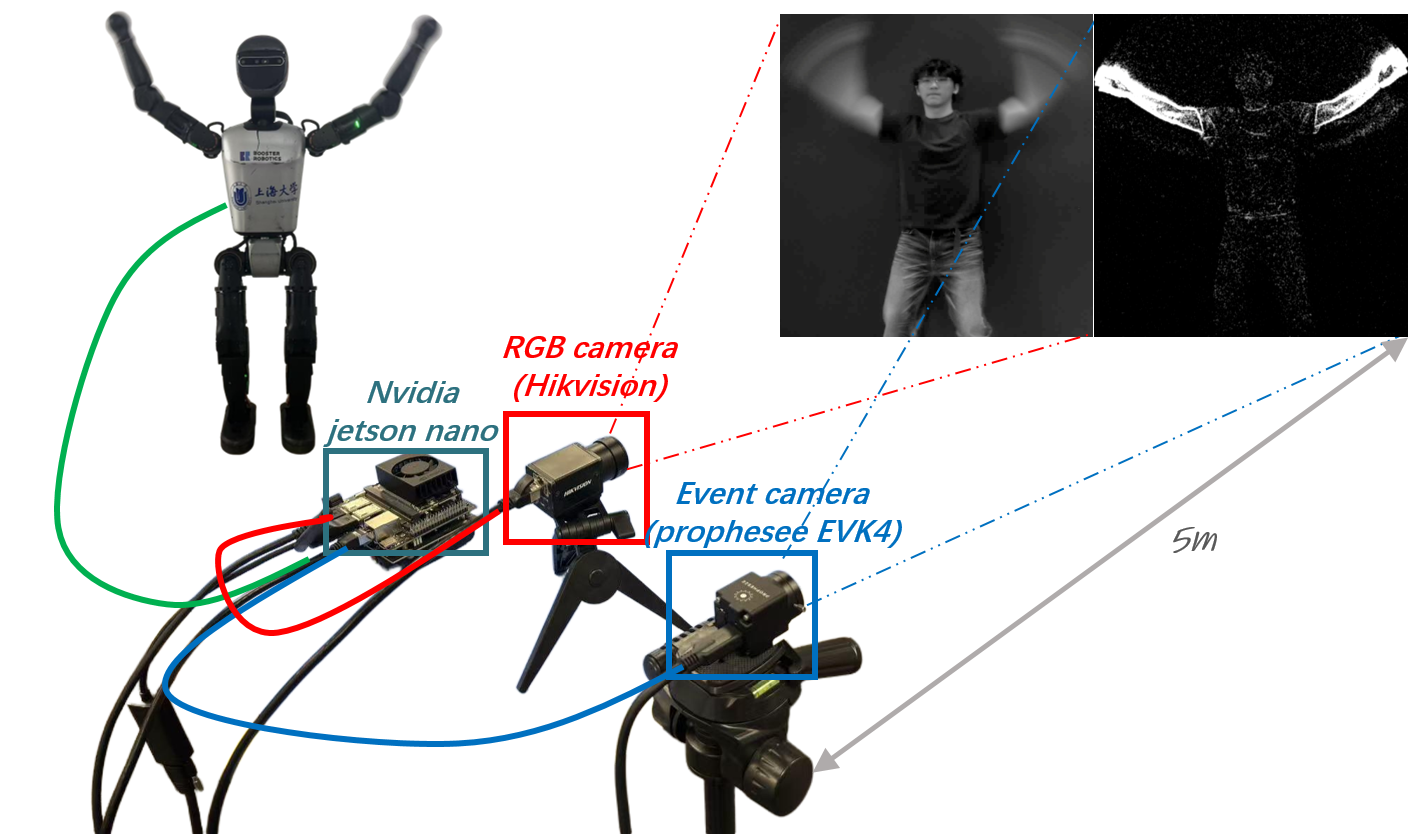}
}{
\fbox{\rule{0pt}{1.2in}\rule{0.9\linewidth}{0pt}}
}
\caption{Experimental setup for the qualitative robot-imitation results in Fig.~\ref{fig:imitation_results}. RGB and event cameras are arranged with similar viewpoints at approximately 5 m from the subject. A Jetson Nano is used to coordinate synchronized dual-camera capture and data transfer under robot-platform power/compute constraints.}
\label{fig:qual_setup}
\end{figure}

\begin{figure}[ht]
\centering
\IfFileExists{figure3.png}{
\includegraphics[width=0.98\linewidth]{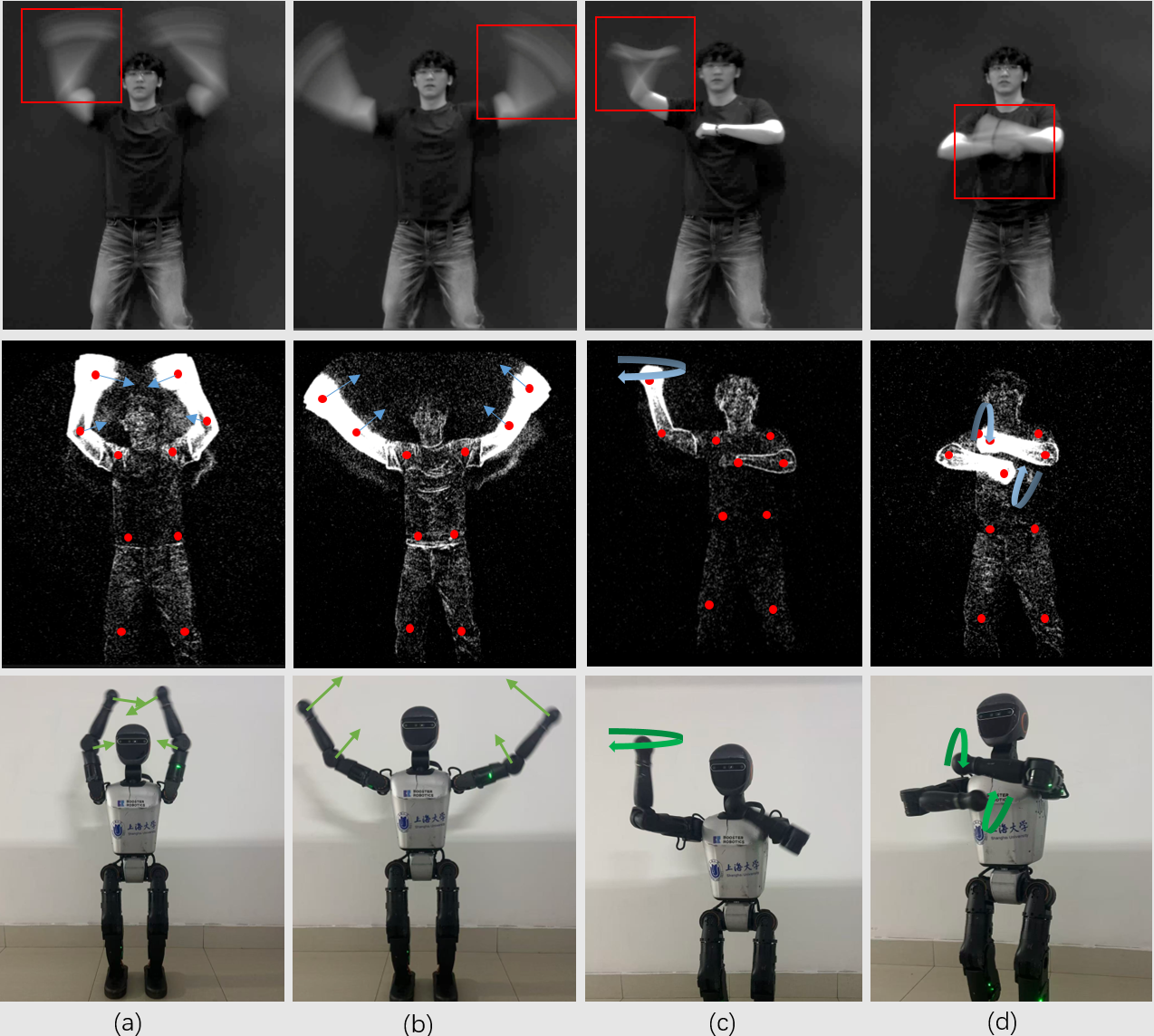}
}{
\fbox{\rule{0pt}{1.4in}\rule{0.95\linewidth}{0pt}}
}
\caption{Qualitative comparison under fast motions with labeled subfigures (a)--(d). Row 1: RGB frames with strong motion blur, where the action becomes hard to identify. Row 2: event frames with clearer motion contours in these examples. Row 3: corresponding event-driven robot imitation results. Arrows mark the dominant hand/arm motion trend, and columns are ordered by capture time. Subfigures (a) and (b) show adjacent waving frames; subfigures (c) and (d) show representative \textit{Gangnam Style} dance frames.}
\label{fig:imitation_results}
\end{figure}

The experimental setup used for the qualitative comparison in Fig.~\ref{fig:imitation_results} is shown in Fig.~\ref{fig:qual_setup}. The RGB camera and the event camera are placed at approximately matched viewpoints, about 5 m from the subject. Due to power and compute limits of the robot-side processing platform, an additional NVIDIA Jetson Nano is used for synchronized dual-camera capture and data transmission.

Figure~\ref{fig:imitation_results} compares RGB observations, event representations, and event-driven robot imitation results for two fast actions.
Subfigures (a) and (b) correspond to adjacent frames of the waving sequence: RGB frames suffer from severe motion blur and the key action pattern becomes difficult to recognize, while event frames preserve clearer motion boundaries in this example.
Subfigures (c) and (d) correspond to representative frames of the \textit{Gangnam Style} dance sequence, where event edges remain well-structured under rapid limb motion and the retargeted robot motion remains temporally coherent.
The arrows in Fig.~\ref{fig:imitation_results} indicate the dominant motion trends of the main actions. During dynamic tasks, the event-based pipeline yields lower Robot Joint Angle RMSE ($4.9^\circ \pm 1.3^\circ$) than RGB ($6.2^\circ \pm 2.0^\circ$), consistent with reduced perceptual jitter and phase lag.

\subsection{Ablation Study}
Replacing the exponential time-surface with a voxel grid increases preprocessing latency from $\approx$2 ms to $>8$ ms. Removing the One-Euro filter raises temporal jitter by about 45\%, and disabling IMU fusion causes up to $15^\circ$ root drift during rapid bending. These results support the selected design choices.

We further observe that disabling confidence-adaptive weighting in TWIST increases limb-endpoint oscillation during partial occlusions, confirming that confidence-aware control coupling is beneficial for stable robot execution. Although this component has limited effect on MPJPE, it notably affects actuator smoothness, which is critical in physical deployment.

\subsection{Discussion}

Since the kinematic structures of humans and humanoid robots differ significantly, 
direct comparison of absolute joint positions in the robot coordinate space is not meaningful. 
Differences in limb lengths, joint limits, and degrees of freedom make Cartesian error metrics 
in centimeters misleading. Therefore, perception accuracy in this work is evaluated in the 
\textit{human skeletal space}, using standard metrics from the human pose estimation literature. 
Robot imitation quality is instead assessed through temporal stability and robustness of tracking.

These results suggest a clear metric trade-off. Under our settings, RGB achieves slightly lower frame-wise pose error in well-lit static conditions, while event sensing provides better timing fidelity and robustness under challenging illumination and rapid motion. For teleoperation, these control-oriented properties often dominate operator experience and safety.

Our optimization target prioritizes \textit{closed-loop controllability} (latency, recovery speed, and command smoothness) over minimal frame-wise pose error, explaining why the event pipeline can improve dynamic interaction quality even when MPJPE is not the best single metric.

\subsection{Threats to Validity and Deployment Notes}
\textbf{Internal validity:} Event and RGB pipelines share the same downstream stack, but sensing principles differ and perfect low-level parity is difficult. We therefore report both perception-space and control-space metrics.

\textbf{External validity:} Our protocol uses upper-body gestures, 12 subjects $\times$ 5 repetitions across four conditions, and one 18-DoF humanoid platform. The paired design improves sensitivity, but these statistics are pilot-scale evidence rather than population-level proof. Clothing/body-size diversity, longer sessions, full locomotion, contact manipulation, and different robot kinematics remain future work.

\textbf{Measurement limitations:} Latency and jitter depend on middleware configuration and hardware clock synchronization. We mitigate this with monotonic timestamps and repeated trials, but cross-lab replication may still shift absolute values.

\textbf{Human-robot safety:} During confidence collapse or prolonged static scenes, the controller falls back to damped hold mode rather than aggressive extrapolation. Safety quantification in this paper focuses on kinematic proxies (recovery delay, post-recovery overshoot, yaw drift, and joint RMSE); contact-force and collision-risk measurements remain future work.

Overall, Tables~\ref{tab:statistical_summary}--\ref{tab:failure_recovery} show that the event pipeline primarily improves robustness and responsiveness rather than absolute MPJPE. Measured perception-front-end power is also lower in our implementation ($3.1\pm0.3$ W vs. RGB $4.0\pm0.4$ W at 30 FPS and $7.8\pm0.6$ W at 120 FPS).

\textbf{System Limitations:} A fundamental limitation of event-based vision is its reliance on relative motion. In completely static scenes where neither the camera nor the subject moves, the event output becomes sparse, increasing re-initialization delay (1.0$\pm$0.3 s in Table~\ref{tab:failure_recovery}). While our system mitigates this through IMU fusion and TWIST regularization, prolonged static periods still benefit from active sensor micro-saccades or complementary RGB/Depth sensing for absolute pose recovery~\cite{b32,b33,b34}. Performance also degrades with very slow motions ($<0.1$ m/s limb tip velocity), low-texture clothing, strong self-occlusion, and dynamic/cluttered backgrounds that generate irrelevant events. IMU integration errors can also accumulate without re-initialization, causing yaw drift (2.4$^\circ$/min over 3 min); periodic yaw reset reduces this to below 0.8$^\circ$/min in our tests.

\section{Conclusion}
\label{sec:conclusion}

We have presented a real-time upper-body motion imitation system for humanoid robots, built around neuromorphic event-based vision. By addressing perception latency, illumination robustness, and kinematic retargeting in a single pipeline, we demonstrate practical advantages in several RGB-challenging scenarios under our setup.

The system achieves millisecond-level responsiveness (23--34 ms) and supports reliable tracking with a sensor dynamic range exceeding 120~dB in our tests. Our findings indicate that the main gain of event sensing in this context is not absolute frame-wise pose accuracy, but robust closed-loop behavior under adverse illumination and rapid motion.

The improvement is not universal across all metrics: RGB remains competitive on frame-wise MPJPE in well-lit static conditions, whereas the event pipeline provides stronger control-oriented robustness.

Future work will address contact-aware manipulation, event+RGB recovery for static scenes, and cross-platform full-body validation.

\section*{Acknowledgment} This work was funded in part by the National Natural Science Foundation of China (62505167), and the Science and Technology Commission of Shanghai Municipality (25ZR1402149).

\begingroup
\let\footnotesize\scriptsize
\bibliographystyle{IEEEtran}
\bibliography{IEEEexample}
\endgroup

\end{document}